%% file: camera_ready.tex
\documentclass[sigconf]{acmart}

\input{math_commands.tex}

\usepackage{algorithm}
\usepackage[noend]{algpseudocode}
\usepackage{amsmath}
\usepackage{booktabs}
\usepackage{multirow}
\usepackage{subcaption}
\usepackage[switch]{lineno}
\newcommand{\mbf}[1]{\mathbf{#1}}
\newcommand\norm[1]{\left\lVert#1\right\rVert}
\AtBeginDocument{%
  \providecommand\BibTeX{{%
    \normalfont B\kern-0.5em{\scshape i\kern-0.25em b}\kern-0.8em\TeX}}}

\copyrightyear{2021}
\acmYear{2021}
\setcopyright{acmcopyright}
\acmConference[KDD '21]{Proceedings of the 27th ACM SIGKDD Conference on Knowledge Discovery and Data Mining}{August 14--18, 2021}{Virtual Event, Singapore}
\acmBooktitle{Proceedings of the 27th ACM SIGKDD Conference on Knowledge Discovery and Data Mining (KDD '21), August 14--18, 2021, Virtual Event, Singapore} \acmPrice{15.00}
\acmDOI{10.1145/3447548.3467386}
\acmISBN{978-1-4503-8332-5/21/08}

\begin{document}
\fancyhead{}

\title[Hard Label Black-box Adversarial Attacks]{Simple and Efficient Hard Label Black-box Adversarial Attacks \\
            in Low Query Budget Regimes}


\author{Satya Narayan Shukla}
 \email{snshukla@cs.umass.edu}
 \affiliation{%
   \institution{
College of Information and Computer Sciences\\
University of Massachusetts Amherst}
\city{Amherst}
\country{USA}
}

 \author{Anit Kumar Sahu}
 \authornote{Work done while at Bosch Center for AI}
 \email{anit.sahu@gmail.com}
 \affiliation{%
   \institution{Amazon Alexa AI}
   \city{Seattle}
   \country{USA}
}

\author{Devin Willmott}
 \email{devin.willmott@us.bosch.com}
 \affiliation{%
   \institution{Bosch Center for AI}
   \city{Pittsburgh}
   \country{USA}
}
\author{J. Zico Kolter}
\email{zkolter@cs.cmu.edu}
 \affiliation{%
   \institution{Computer Science Department\\
   Carnegie Mellon University,\\Bosch Center for AI}
   \city{Pittsburgh}
   \country{USA}
}

\renewcommand{\shortauthors}{Shukla et al.}

\begin{abstract}
We focus on the problem of black-box adversarial attacks, where the aim is to generate adversarial examples for deep learning models solely based on information limited to output label~(hard label) to a queried data input. 
We propose a simple and efficient Bayesian Optimization~(BO) based approach for developing black-box adversarial attacks. Issues with BO's performance in high dimensions are avoided by searching for adversarial examples in a structured low-dimensional subspace. We demonstrate the efficacy of our proposed attack method by evaluating both $\ell_\infty$ and $\ell_2$ norm constrained untargeted and targeted hard label black-box attacks on three standard datasets - MNIST, CIFAR-10 and ImageNet. Our proposed approach consistently achieves $2\times \text{ to } 10\times$ higher attack success rate while 
requiring $10\times$ to $20 \times$ fewer queries compared to the current state of the art black-box adversarial attacks.\footnote{
Implementation available at: \url{https://github.com/satyanshukla/bayes_attack}}
\end{abstract}



\maketitle
\input{intro}
\input{related}

\input{model}

\input{experiments2}

\input{conclusions}

\bibliographystyle{ACM-Reference-Format}
\bibliography{bibliography}

\end{document}

%% file: math_commands.tex
\usepackage{amsmath,amsfonts,bm}

\def\eqref#1{equation~\ref{#1}}

\def\1{\bm{1}}

\DeclareMathAlphabet{\mathsfit}{\encodingdefault}{\sfdefault}{m}{sl}
\SetMathAlphabet{\mathsfit}{bold}{\encodingdefault}{\sfdefault}{bx}{n}

\DeclareMathOperator*{\argmax}{arg\,max}
\DeclareMathOperator*{\argmin}{arg\,min}

%% file: intro.tex
\section{Introduction}
\label{sec:intro}


Neural networks are now well-known to be vulnerable to \textit{adversarial examples}: additive perturbations that, when applied to the input, change the network's output classification \citep{goodfellow2014explaining}. Work investigating this lack of robustness to adversarial examples often takes the form of a back-and-forth between newly proposed \textit{adversarial attacks}, methods for quickly and efficiently crafting adversarial examples, and corresponding defenses that modify the classifier at either training or test time to improve robustness.
The most successful adversarial attacks use gradient-based optimization methods \citep{goodfellow2014explaining,madry2017towards}, which require complete knowledge of the architecture and parameters of the target network; this assumption is referred to as the \textit{white-box} attack setting. Conversely, the more realistic \textit{black-box} setting requires an attacker to find an adversarial perturbation without such knowledge: information about the network can be obtained only through querying the target network, i.e., supplying an input to the network and receiving the corresponding output. In terms of information obtained from queries, it can be further categorized into \emph{soft-label} and \emph{hard-label}. As far as the soft-label information is concerned, the information for a query is typically in terms of the logit values or the evaluation of the loss function at that particular input. The more realistic and challenging of the two, i.e., hard-label information obtained from a query is just the label of the input fed into the network.

In real-world scenarios, it is extremely improbable for an attacker to have unlimited bandwidth to query a target classifier. In the evaluation of black-box attacks, this constraint is usually formalized via the introduction of a \textit{query budget}: a maximum number of queries allowed to query the model per input, after which an attack is considered to be unsuccessful. Several recent papers have proposed attacks specifically to operate in this query-limited context \citep{bandits,nes,zoo,autozoom,Chen2018fw,parsimonious,cheng2018query, signopt}; nevertheless, these papers typically consider query budgets in the order of 10,000 or 100,000. This leaves open questions as to whether black-box attacks can successfully attack a deep learning based classifier in severely query limited settings, e.g., with query budgets below 1000. Furthermore, restricting the information available from querying to be hard-label only, makes the aforementioned direction even more challenging. In such a query-limited regime, it is natural for an attacker to use the entire query budget, so we ask the pertinent question: \emph{In a constrained query limited setting, can one design query efficient yet successful black-box adversarial attacks, where the queried information is restricted to being hard-label?}

This work proposes a simple and efficient hard-label black-box attack method grounded in Bayesian optimization \citep{bayesopt,bayesopt_tutorial}, which has emerged as a state-of-the-art black-box optimization technique in settings where minimizing the number of queries is of paramount importance. Straightforward application of Bayesian optimization to the problem of finding adversarial examples is not feasible: the input dimension of even a small neural network-based image classifier is orders of magnitude larger than the standard use case for Bayesian optimization. Rather, we show that we can bridge this gap by performing Bayesian optimization in a reduced-dimension setting by considering structured subspaces and mapping it back to full input space to obtain our final perturbation. We explore several mapping techniques and find that reducing the search space to a structured subspace composed of Fast Fourier Transform~(FFT) basis vectors and a simple nearest-neighbor upsampling method allows us to sufficiently reduce the optimization problem dimension such that Bayesian optimization can find adversarial perturbations with more success than existing hard-label black-box attacks in query-constrained settings.

We compare the efficacy of our adversarial attack with a set of experiments attacking several models on MNIST \citep{mnist}, CIFAR10 \citep{cifar} and ImageNet \citep{imagenet}. We perform both $\ell_\infty$ and $\ell_2$ norm constrained black-box attacks. We consider both untargeted and targeted attack settings. Results from these experiments show that with small query budgets up to $1000$, the proposed Bayes attack achieves significantly better attack success rates than those of existing methods, and does so with far smaller average query counts.  Furthermore, ablation experiments are performed to compare the effectiveness of the configurations considered in our attack setup, i.e., selection of the structured low dimensional subspace and mapping techniques to generate adversarial perturbations in the original image space.  Given these results, we argue that, despite being a simple approach (indeed, largely \emph{because} BO is such a simple and standard approach for black-box optimization), Bayes attack should be a standard baseline for any hard-label black-box adversarial attack task in the future, especially in the small query budget regime.

%% file: related.tex
\section{Related Work}
Within the black-box setting, adversarial attacks can be further categorized by the exact nature of the information received from a query. 
This work exists in the restrictive \textit{hard-label} or \textit{decision-based} setting, where queries to the network yield only the predicted class of the input, with no other information about the network's final layer output. 
The most successful work in this area is OPT attack \citep{cheng2018query}, which reformulates the problem as a search for the direction of the nearest decision boundary and employs a random gradient-free method, and Sign-OPT attack \citep{signopt}, which refines this approach by replacing binary searches with optimization via sign SGD. Building upon previous work, \citet{rays2020} propose to find the closest decision boundary by directly searching along a discrete set of ray directions. Although, this approach is only suitable for generating $\ell_\infty$ norm attacks.
In Boundary Attack \citep{brendel2017decision}, attacks are generated via random walks along the decision boundary with rejection sampling. Several other attacks have extended this work and refined its query efficiency: HopSkipJumpAttack \citep{Chen2019HopSkipJumpAttackAQ} does so with improved gradient estimation while Guessing Smart \citep{brunner2019guessing} incorporates low-frequency perturbations, region masking, and gradients from a surrogate model. In both cases, significant issues remain: the former still requires queries numbering above 10,000 to produce adversarial examples with small perturbations, and the latter relies on resources required to train a surrogate model.

Most work in black-box adversarial attacks has been dedicated to \textit{score-based} or \textit{soft label} attacks, where queries return the entire output layer of the network, either as logits or probabilities.
Relative to hard-label attacks queries in the soft label setting receive a large amount of information from each query, making them amenable to approaches from a wide variety of optimization fields and techniques.
The most popular avenue is maximizing the network loss with zeroth-order methods via derivative-free gradient estimation, such as those proposed in \citet{bandits}, which uses time-dependent and data-dependent priors to improve the estimate, as well as \citet{nes}, which estimates gradients using natural evolution strategies (NES).
Other methods search for the best perturbation outside of this paradigm; \citet{parsimonious} cast the problem of finding an adversarial perturbation as a discrete optimization problem and use local search methods to solve it.
These works all search for adversarial perturbations within a search space with a hard constraint on perturbation size; others \citep{zoo, autozoom,guo2019simple} incorporate a soft version of this constraint and perform coordinate descent or random walks to decrease the perturbation size while ensuring the perturbed image is misclassified.

A separate class of \textit{transfer-based} attacks trains a second, fully-observable substitute network, attack this network with white-box methods, and transfer these attacks to the original target network.
These may fall into one of the preceding categories or exist outside of the distinction: in \citet{papernot2017practical}, the substitute model is built with score-based queries to the target network, whereas \citet{liu2016delving} train an ensemble of models without directly querying the target network at all.
These methods come with their drawbacks: they require training a substitute model, which may be costly or time-consuming; attack success is frequently dependent on similarity between substitute and target networks, and overall attack success tends to be lower than that of gradient-based methods.

Beyond these categories, we note that our method here sits among several recent works that find improved success rates and query efficiency from restricting their search for adversarial perturbations to particular low-dimensional subspaces. One common approach is to generate adversarial perturbations from low-frequency noise, such as in \citet{lowfreq}, which improves existing attacks \citep{nes,brendel2017decision} by confining searches to subspaces of low-frequency basis vectors, and in \citet{brunner2019guessing}, which employs a Perlin noise basis. In a similar vein, \citet{bandits} exhibit that local spatial similarity of images, i.e., the tendency of nearby pixels to be similar, extends to gradients, and uses this observation to motivate focusing on tile-based perturbations.

Finally, there has been some recent interest in leveraging Bayesian optimization (BO) for constructing adversarial perturbations.
For example, \citet{Zhao2019} use BO to solve the $\delta$-step of an alternating direction of method multipliers approach, \citet{procedural} searches within a set of procedural noise perturbations using BO and \citet{alg_assurance} use BO to find maximal distortion error by optimizing perturbations defined using $3$ parameters.
On the other hand, prior work in which Bayesian optimization plays a central role, the use cases and experiments are performed only in relatively low-dimensional settings, highlighting the main challenge of its application: \citet{suya2017query} examine an attack on a spam email classifier with 57 input features, and in \citet{munoz2017bayesian} image classifiers are attacked but notably, the attack does not scale beyond MNIST classifiers. More importantly, these prior works along with \citet{bayesopt_adv} focus only on score-based setting and it is not clear if such methods would work in a more restricted hard-label or decision-based setting.
In contrast to these past works, the main contribution of this paper is to show that Bayesian Optimization presents as a \emph{scalable, query-efficient} alternative for large-scale hard-label black-box adversarial attacks when combined with searching in structured low dimensional subspaces and employing mapping techniques to get the final adversarial perturbation.  To the best of our knowledge, this is the first paper considering hard label black-box attacks with query budgets under $1000$.

%% file: model.tex
\section{Problem Formulation}
\label{sec:problem}

The following notation and definitions will be used throughout the remainder of the paper.
Let $F$ be the target neural network. 
We assume that $F:\mathbb{R}^{d'} \rightarrow \{1, \ldots K\}$ is a $K$-class image classifier that takes normalized RGB inputs:$\mathbf{x} \in \mathbb{R}^{d'}$ where each channel of each pixel is bounded between $0$ and $1$, $y \in \{1, \ldots, K\}$ denotes the original label, and the corresponding output $F(\mathbf{x})$ is the final output label or class.
\begin{align}
\label{eq:obj}
    & \max_{\bm{\delta}} \, f(\mbf{x}, y, \bm{\delta})\\
    & \text{subject to} \;\; \Vert \bm{\delta} \Vert_p \leq \epsilon \;\; \text{and} \;\;  (\mbf{x} + \bm{\delta}) \in [0, 1]^{d'},  \nonumber\\
 & \text{where} \hspace{0.1in} f(\mbf{x}, y, \bm{\delta}) = \begin{cases}
0 & \text{if } F(\mbf{x} + \bm{\delta}) \neq y \\
-1 & \text{if } F(\mbf{x} + \bm{\delta}) = y
\end{cases} \nonumber
\end{align}
Rigorous evaluation of an adversarial attack requires careful definition of a \textit{threat model}: a set of formal assumptions about the goals, knowledge, and capabilities of an attacker \citep{carlini2017}. We assume that, given a correctly classified input image $\mathbf{x}$, the goal of the attacker is to find a
perturbation $\boldsymbol{\delta}$ such that $\mathbf{x}+\boldsymbol{\delta}$ is misclassified, i.e., \mbox{$ F(\mbf{x} + \bm{\delta}) \neq F(\mbf{x})$}.
We operate in the hard-label black-box setting, where we have no knowledge of the internal workings of the network, and a query to the network $F$ yields only the final output class (top-1 prediction). 
To enforce the notion that the adversarial perturbation should be small, we take the common approach of requiring that $\|\boldsymbol{\delta}\|_p$ be smaller than a given threshold $\epsilon$ in some $\ell_p$ norm. 
In this work, we specifically focus on $\ell_2$ and $\ell_\infty$ norms. 
Finally, we let $t$ denote the query budget, i.e., if an adversarial example is not found after $t$ queries to the target network, the attack is considered to be unsuccessful.
In line with most adversarial attack setups, we pose the attack as a constrained optimization problem, defined in Equation \ref{eq:obj}.
%
%
Crucially, the input $\mathbf{x}\, + \, \boldsymbol{\delta}$ to $f$ is an adversarial example for $F$ if and only if $f(\mathbf{x}, y, \boldsymbol{\delta}) = 0$.
Though our objective function is straightforward, we empirically show that it leads to significant performance improvements over the current state of the art for hard-label black-box attacks for both $\ell_\infty$ and $\ell_2$ threat models. 

We briefly note that the above threat model and objective function were chosen for simplicity and for ease of directly comparing with other black-box attacks, but the proposed method is compatible with many other threat models. For example, we may change the goals of the attacker from untargeted to the targeted setting or measure $\boldsymbol{\delta}$ in $\ell_1$ norm instead of $\ell_2$ and $\ell_{\infty}$ norms with appropriate modifications to the objective function and constraints in \eqref{eq:obj}.



\begin{figure*}[t]
\centering
\includegraphics[width=0.7\linewidth]{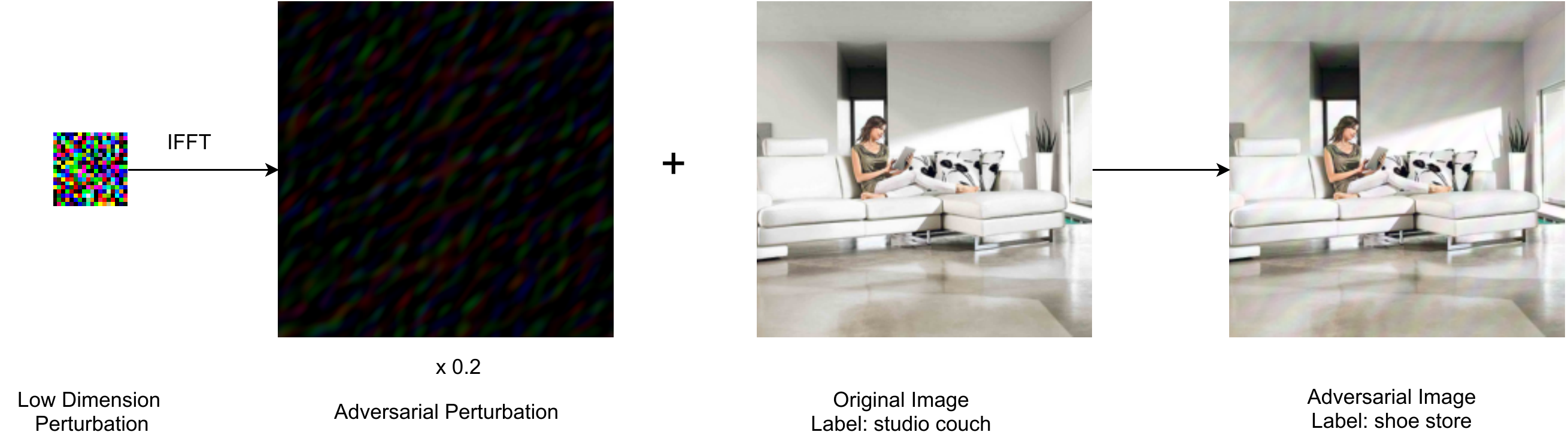}
\caption{An illustration of a black-box adversarial attack performed by the proposed {Bayes attack} on {ResNet50} trained on ImageNet. Images from the left: first figure shows learned perturbation in low dimension $d' = 972 (3\times18\times18)$; second figure is the final adversarial perturbation $(3\times224\times224)$  obtained by using inverse FFT; third figure is the original image (input size is $3\times 224 \times 224$) which is initially classified as {\it studio couch}; last image is the final adversarial image obtained by adding the adversarial perturbation to the original image. {ResNet50} classifies the final adversarial image as {\it shoe store} with high probability.}   
\label{fig:ex}
\end{figure*}

\section{Proposed Attack Method}
In this section, we present the proposed method for solving the optimization problem in \eqref{eq:obj}. 
We begin with a brief description of Bayesian optimization \citep{bayesopt} followed by its application to generating black-box adversarial examples. Finally, we describe our method for attacking a classifier with a high input dimension (e.g. ImageNet) in a query-efficient manner.
\begin{algorithm*}[t]
\caption{Black-box Adversarial Attack using Bayesian Optimization}
\label{alg:bayesopt}
\begin{algorithmic}[1]
\Procedure{\textsc{Bayes-Attack}}{$x_0, y_0$}
    \State $\mathcal{D}=\{(\bm{\delta}_1, v_1), \cdots, (\bm{\delta}_{n_0}, v_{n_0})\}$   \Comment{Quering randomly chosen $n_0$ points.}
    \State Update the GP on $\mathcal{D}$ \Comment{Updating posterior distribution using available points}
    \State $t \leftarrow n_0$ \Comment{Updating number of queries till now}
    \While{$t \leq T$}
    \State $\bm{\delta}_t \leftarrow \argmax_{\bm{\delta}} \mathcal{A}(\bm{\delta} \;|\; \mathcal{D})$ \Comment{ Optimizing the acquisition function over the {GP}}
    \State $\bm{\delta}_t \leftarrow \Pi^p_{B(\mbf{0}, \epsilon)}(\bm{\delta}_t)$ \Comment{Projecting perturbation on $\ell_p$-ball} 
    \State $\bm{\delta}_t \leftarrow \; \text{map}(\bm{\delta}_t)$  \Comment{Mapping perturbation from low dimension subspace to full input space}
    \State $v_t \leftarrow f(\mbf{x}_0, y_0, \bm{\delta}_t)$     \Comment{Querying the model}
    \State $t \leftarrow t + 1$
    \If{$v_t < 0$}
    \State $\mathcal{D} \leftarrow \mathcal{D} \cup (\bm{\delta}_t, v_t)$ and update the GP \Comment{Updating posterior distribution}
    \Else 
    \State \Return $\bm{\delta}_t$ \Comment{Adversarial attack successful}
    \EndIf
    \EndWhile
    \State \Return $\bm{\delta}_t$ \Comment{Adversarial attack unsuccessful}
\EndProcedure
\end{algorithmic}
\end{algorithm*}

\subsection{Bayesian Optimization}

Bayesian optimization (BO) is a method for black-box optimization particularly suited to problems with low dimension and expensive queries. 
Bayesian optimization consists of two main components: a Bayesian statistical model and an acquisition function. The Bayesian statistical model, also referred to as the surrogate model, is used for approximating the objective function: it provides a Bayesian posterior probability distribution that describes potential values for the objective function at any candidate point. This posterior distribution is updated each time we query the objective function at a new point. The most common surrogate model for Bayesian optimization is Gaussian processes (GPs) \citep{gp}, which define a prior over functions that are cheap to evaluate and are updated as and when new information from queries becomes available.
We model the objective function $h$ using a GP with prior distribution $\mathcal{N}(\mu_0, \Sigma_0)$ with constant mean function $\mu_0$ and Matern kernel \citep{matern,practical} as the covariance function $\Sigma_0$, which is defined as:
\begin{align*}
    \Sigma_0 (\mbf{x}, \mbf{x}') &= \theta_0^2 \exp({-\sqrt{5} r}) \bigg(1 +  \sqrt{5} r + \frac{5}{3} r^2\bigg), \;\;\;\;\;\;\;\;\;\;\;
     \nonumber
\end{align*}
where $r^2 = \sum_{i=1}^{d'} \frac{(x_i - x'_i)^2}{\theta_i^2}$, $d'$ is the dimension of input and $\{\theta_i\}_{i=0}^{d'}$ and $\mu_0$ are hyperparameters. We select hyperparameters that maximize the posterior of the observations under a prior \citep{matern, bayesopt_tutorial}.

The second component, the acquisition function $\mathcal{A}$, assigns a value to each point that represents the utility of querying the model at this point given the surrogate model.
We sample the objective function $h$ at $\mbf{x}_n = \argmax_\mathbf{x} \mathcal{ A}(\mathbf{x} | \mathcal{D}_{1 : n-1})$ where $\mathcal{D}_{1 : n-1}$ comprises of $n-1$ samples drawn from $h$ so far.  Although this itself may be a hard (non-convex) optimization problem to solve, in practice we use a standard approach and approximately optimize this objective using the LBFGS algorithm.  There are several popular choices of acquisition function; we use expected improvement (EI) \citep{bayesopt}, which is defined by $\text{EI}_n (\mathbf{x}) = \mathbb{E}_n \left[\max \; \left(h(\mathbf{x}) - h_n^*, 0\right)\right],$ where $\mathbb{E}_n [\cdot] = \mathbb{E} [\cdot | \mathcal{D}_{1 : n-1}]$  denotes the expectation taken over the posterior distribution given evaluations of $h$ at $\mathbf{x}_1, \cdots, \mathbf{x}_{n-1}$, and $h_n^*$ is the best value observed so far.
%
%

Bayesian optimization framework as shown in Algorithm \ref{alg:bayesopt} runs these two steps iteratively for the given budget of function evaluations. It updates the posterior probability distribution on the objective function using all the available data. Then, it finds the next sampling point by optimizing the acquisition function over the current posterior distribution of GP. The objective function $h$ is evaluated at this chosen point and the whole process repeats. 

In theory, we may apply Bayesian optimization directly to the optimization problem in \eqref{eq:obj} to obtain an adversarial example, stopping once we find a point where the objective function reaches $0$. In practice, Bayesian optimization's speed and overall performance fall dramatically as the input dimension of the problem increases. This makes running Bayesian optimization over high dimensional inputs such as ImageNet (input dimension $3 \times 299 \times 299 = 268203$) practically infeasible; we therefore require a method for reducing the dimension of this optimization problem.

\subsection{Bayes Attack: Generating Adversarial Examples using Bayesian Optimization}

Black-box attack methods tend to require a lot of queries because of the search space being high dimensional. The query complexity of these methods depends on the adversarial subspace dimension compared to the original image space. These methods can be improved by finding a structured low dimensional subspace so as to make the black-box optimization problem feasible and thereby resulting in adversarial examples with fewer queries and higher success rates.
We define two low dimension subspaces favorable for generating $\ell_2$ and $\ell_\infty$ norm constrained hard label black-box attacks. 
\subsubsection*{\bf Low Dimensional Subspace for $\ell_2$ norm constrained attack}
To generate $\ell_2$ norm constrained adversarial examples, our attack method utilizes low-frequency fast Fourier transform (FFT) basis vectors. FFT is a linear transform that, when applied to a natural image, results in a representation in frequency space by sine and cosine basis vectors. For a given image $\mathbf{x} \in \mathbb{R}^{d\times d}$, the output of the FFT transform $\mbf{X} := \text{FFT}(\mbf{x})$ is defined by
\begin{align}
    \mbf{X}[u, v] =& \frac{1}{d} \sum_{i = 0}^{d-1} \sum_{j = 0}^{d-1} \mbf{x}[i , j] \exp\bigg[{-j\frac{2\pi}{d}(u\cdot i + v\cdot j)}\bigg] 
\end{align}
where $\frac{1}{d}$ is the normalization constant to obtain isometric transformation, i.e., $\norm{\mbf{x}}_2 = \norm{\text{FFT}(\mbf{x})}_2$. The inverse fast fourier transform $\mbf{x} = \text{IFFT}(\mbf{X})$ is defined by:
\begin{align}
    \mbf{x}[i, j] =& \frac{1}{d}\sum_{u = 0}^{d-1} \sum_{v = 0}^{d-1} \mbf{X}[u , v]
    \exp\bigg[{j\frac{2\pi}{d}(u\cdot i + v\cdot j)}\bigg]
\end{align}
The isometric property holds in reverse direction too, i.e. $\norm{\mbf{X}}_2 = \norm{\text{IFFT}(\mbf{X})}_2$. 
 For multi-channel (colored) images, both FFT and IFFT can be applied channel-wise independently. The low-frequency cosine and sine basis vectors are represented by small $u,v$ values in real and complex components of $\mbf{X}(u, v)$ respectively. To restrict to low-frequencies, we allow only elements in the top-left $\lfloor rd \rfloor \times \lfloor rd\rfloor$ square of $\mbf{X}$ to have nonzero entries, where $r \in (0, 1]$ is a parameter that controls how large we allow our search space to be; that is, we enforce $\mbf{X}(u, v) = 0$ if $u > \lfloor rd \rfloor$ or $v >\lfloor rd\rfloor$.
 The adversarial perturbation is then obtained by computing $\text{IFFT}(\mbf{X})$.
 
 To further reduce the dimension of this search space, we may also omit all sine or all cosine FFT basis vectors by respectively constraining the real or imaginary parts of $\mbf{X}$ to be zero. An ablation study exploring the effect of removing sine or cosine FFT basis vectors on performance is shown in  Section \ref{sec:basis}.

\subsubsection*{\bf Low Resolution Subspace for $\ell_\infty$ norm constrained attack}
Our method uses a data dependent prior to reduce the search dimension of the perturbation for generating $\ell_\infty$ norm constrained adversarial examples. We empirically show that we can utilize the property of spatial local similarity in images to generate adversarial perturbations. For a given image $\mbf{x} \in \mathbb{R}^{d\times d}$, we search for perturbations in a lower resolution image space $\mbf{X} \in \mathbb{R}^{\lfloor rd \rfloor \times \lfloor rd \rfloor}$ where $r \in (0, 1]$ and use nearest neighbor interpolation~(NNI) $\mbf{x}' = \text{NNI}(\mbf{X})$ to obtain the final adversarial perturbation. We note that $\mbf{x}' \neq \mbf{x}$. The NNI transformation leads to equivalent $\ell_\infty$ norms, i.e., $\norm{\mbf{X}}_\infty = \norm{\text{NNI}(\mbf{X})}_\infty$.  
For multi-channel images, NNI can be applied channel-wise independently.

\subsubsection*{\bf Searching in Low Dimension Subspace}
We use Bayesian optimization to search for the perturbation in low dimension subspace $(\lfloor rd \rfloor \times \lfloor rd\rfloor)$ where $r \in (0,1]$ and then use the relevant mapping (IFFT for $\ell_2$ or NNI for $\ell_\infty$) to obtain the adversarial perturbation in the original image space. This helps in reducing the query complexity for our attack framework, due to the reduction of the search space to a structured low dimensional subspace.

We let $\Pi^{p}_{B(\mbf{0}, \epsilon)}$ be the projection onto the $\ell_p$ ball of radius $\epsilon$ centered at the origin. Our method maps the learned perturbation in low dimension subspace back to the original input space to obtain the adversarial perturbation. We maintain the $\ell_p$ norm constraint by projecting the low dimension subspace perturbation on a $\ell_p$ ball of radius $\epsilon$. Since our mapping techniques (IFFT for $\ell_2$ and NNI for $\ell_\infty$) do not change their respective norms, the final adversarial perturbation obtained after mapping to original input space also follows the $\ell_p$ constraint.  

We describe the proposed attack method in Algorithm \ref{alg:bayesopt} where $\mbf{x}_0 \in \mathbb{R}^{d\times d}$ and $y_0 \in \{1 , \ldots, K\}$ denote the original input image\footnote{For simplicity, we assume a 2D image here, our method can be easily applied to multi-channel images.} and label respectively. The goal is to learn an adversarial perturbation $\bm{\delta} \in \mathbb{R}^{\lfloor rd \rfloor \times \lfloor rd\rfloor}$ in a much lower dimension, i.e., $r << 1$. 
We begin with a small dataset $\mathcal{D} = \{(\bm{\delta}_1, v_1), \cdots, (\bm{\delta}_{n_0}, v_{n_0})\}$ where each $\bm{\delta}_n$ is a $\lfloor rd \rfloor \times \lfloor rd\rfloor$ perturbation sampled from a given distribution and $v_n$ is the function evaluation at $\bm{\delta}_n$.
We iteratively update the posterior distribution of the GP using all available data and query new perturbations obtained by maximizing the acquisition function over the current posterior distribution of GP until we find an adversarial perturbation or run out of query budget. The Bayesian optimization iterations run in low dimension subspace $\lfloor rd \rfloor \times \lfloor rd\rfloor$ but for querying the model we project and map to the original image space and then add the perturbation to the original image to get the perturbed image to conform to the input space of the model. To generate a successful adversarial perturbation, it is necessary and sufficient to have $v_t \geq 0$, as described in Section \ref{sec:problem}. We call our attack successful with $t$ queries to the model if the Bayesian optimization loop exits after $t$ iterations (line 14 in Algorithm \ref{alg:bayesopt}), otherwise it is unsuccessful.  The final adversarial image can be obtained by mapping the learned perturbation back to the original image space and adding it to the original image as shown in Figure \ref{fig:ex}. For multi-channel image, the low dimension subspace is of form $3 \times \lfloor rd \rfloor \times \lfloor rd\rfloor$ and the whole algorithm works the same way. In this work, we focus on $\ell_\infty$ and $\ell_2$-norm perturbations, where the respective projections for a given perturbation bound $\epsilon$ are defined by $\Pi^{\infty}_{B(\mbf{x_0}, \epsilon)}(\mbf{x})= 
\min\left\{ \max \{\mbf{x_0} - \epsilon, \mbf{x} \}, \mbf{x_0} + \epsilon\right\},$ and  $\Pi^{2}_{B(\mbf{x_0}, \epsilon)}(\mbf{x}) = \argmin_{\norm{y-x_0}_2 \leq \epsilon} \norm{y-x}_2$. The initial choice of the dataset $\mathcal{D}$ to form a prior can be done using the standard normal distribution, uniform distribution, or even in a deterministic manner (e.g. with Sobol sequences). 

Finally, we note that even though our simple objective leads to the same objective values at queried points in the initial search phase, the Gaussian process provides an expression for the uncertainty at each point in the feasible space, and updates this uncertainty as each new data point is queried. This information is taken into account in the acquisition function, so that, for instance, Bayesian optimization will not query a point that is close to points that have already been queried. 

%% file: experiments2.tex
\section{Experiments}
\begin{table*}[ht]
 \centering
 \caption{Results for $\ell_\infty$ untargeted attacks on ImageNet classifiers with a query budget of $1000$.}
\begin{tabular}{lcc| cc| cc}
\toprule
    & \multicolumn{2}{c}{ResNet50}                                 & \multicolumn{2}{c}{Inception-v3}        & \multicolumn{2}{c}{VGG16-bn}  \\
{\bf Method}               & {success} & {avg. query} & {success} & {avg. query} & {success} & {avg. query} \\
\midrule
OPT attack           & 5.73                        & 246.31                         & 2.87                        & 332.17                         & 7.53                        & 251.21                         \\
Sign-OPT             & 10.31                       & 660.37                         & 7.51                        & 706.3                          & 15.85                       & 666.87                         \\
\textbf{Bayes attack}    & \textbf{67.48}              & \textbf{45.94}                 & \textbf{44.29}              & \textbf{72.31}                 & \textbf{78.47}              & \textbf{33.7}   \\
\bottomrule
\end{tabular}

\label{table:query_compare_linf_imagenet}
\end{table*}

\begin{table*}[t]
\centering
\caption{Results for $\ell_\infty$ untargeted and targeted attacks on MNIST and CIFAR-10 with a query budget of $1000$.}
\begin{tabular}{l c c c c | c c  c c }
\toprule
{\bf Method} &  \multicolumn{4}{c}{\bf Untargeted} & \multicolumn{4}{c}{\bf Targeted} \\
\midrule
 & \multicolumn{2}{c}{\bf MNIST} & \multicolumn{2}{c}{\bf CIFAR-10} & \multicolumn{2}{c}{\bf MNIST} & \multicolumn{2}{c}{\bf CIFAR-10} \\
 & {success} & {avg. query} & {success} & {avg. query} & {success} & {avg. query} & {success} & {avg. query}\\
\midrule
OPT attack & 2.91 & 657.93 & 12.55 & 271.24  & 0.0 & $-$ & 0.0 &  $-$\\
Sign-OPT &  7.02 & 682.36 &  31.87 & 679.39  &  2.41 & 975.67 &  3.50 & 937.65\\
\textbf{Bayes attack} & \bf{90.35} & \bf{27.56} & \bf{70.38} & \bf{75.88} & \bf{26.23} & \bf{130.03} & \bf{48.93} & \bf{149.15}\\
\bottomrule
\end{tabular}

\label{table:query_compare_linf_mnist_cifar}
\end{table*}





Our experiments focus on both untargeted and targeted attack settings. Given an image correctly classified by the model, in untargeted attack setting the goal is to produce an $\ell_p$-constrained perturbation such that applying this perturbation to the original image causes misclassification while in targeted attack setting the goal is to produce perturbation such that the resulting image classifies to a given target class. We evaluate both $\ell_\infty$ and $\ell_2$ norm constrained hard label black-box attacks on three different standard datasets - ImageNet \citep{imagenet}, CIFAR-10 \citep{cifar}, and MNIST \citep{mnist}. For a fair comparison with previous works, we use the same networks used in previous hard-label attack methods. For MNIST and CIFAR-10, we use CNN architectures with 4 convolution layers, 2 max-pooling layers, and 2 fully-connected layers. Using the parameters provided in \citep{cheng2018query,signopt}, we achieved accuracies of $99.5\%$ and $82.5\%$ on MNIST and CIFAR-10 respectively, as reported. For ImageNet, we attack the pre-trained models $-$ ResNet50 \citep{resnet}, Inception-v3 \citep{inception} and VGG16-bn \citep{vgg}, similar to the existing methods.
We also perform ablation studies on ImageNet by exploring different low dimension structured subspaces and examining different upsampling techniques.

To compare performance, we randomly selected a subset of 1000 images, normalized to $[0,1]$ from the test set in the case of CIFAR-10 and MNIST and validation set in the case of ImageNet. We primarily compare the performance of the {Bayes attack} with that of other hard-label black-box attacks for small query budgets and report success rates and average queries. For the baseline methods, we use the implementations made available by the authors and use hyperparameters as suggested in their respective papers. We define success rate as the ratio of the number of images successfully perturbed for a given query budget to the total number of input images. In all experiments, images that are already misclassified by the target network are excluded from the set; only images that are initially classified with the correct label are attacked. For each method of attack and each target network, we compute the success rate and the average number of queries used to attack among images that were successfully perturbed.


\subsection{Empirical Protocols}
We treat the size of the low dimensional subspace used for running the Bayesian optimization loop as a hyperparameter. 
For both $\ell_2$ and $\ell_\infty$ attacks, we tune the low dimensional subspace size $\lfloor rd \rfloor \times \lfloor rd \rfloor $ over the range $ rd \in [5, 18]$. For $\ell_2$ attacks, we treat the option of representing the subspace with sine and cosine FFT basis vectors separately or together as a hyper-parameter.
We initialize the GP with $n_0 = 5$ samples drawn from a standard normal distribution in case of $\ell_\infty$ attacks, and the uniform distribution $[-1, 1]$ for $\ell_2$ attacks. 
The intuition behind initialization is that for the $\ell_2$ model, by operating in low dimension subspace, we enforce most of the frequency components to be zero, hence we initialize the rest of
the components with uniform distribution. While for $\ell_\infty$ model, initialization with standard normal would mean that
most values in our perturbations are zero or close to zero which would be more imperceptible than uniformly initialized
values. For all the experiments in this section, we use expected improvement as the acquisition function. We also examined other acquisition functions (posterior mean, probability of improvement, upper confidence bound) and observed that our method works equally well with other acquisition functions. 
We independently tune the hyperparameters on a small validation set and exclude it from our final set of images to be attacked. We used BoTorch\footnote{\url{https://botorch.org/}} packages for implementation.

\subsection{Untargeted $\ell_\infty$ Attacks}
First, we compare the performance of the {Bayes attack} against OPT \citep{cheng2018query} and {Sign-OPT} \citep{signopt}, which is the current state of the art among hard-label black-box attacks within the $\ell_\infty$ threat model. 
We evaluate the performance of all the methods with a budget of up to $1000$ queries. We set $\ell_\infty$ perturbation bound $\epsilon$ to $0.05$ for CIFAR-10 and ImageNet, and $0.3$ for MNIST. Table \ref{table:query_compare_linf_imagenet} and  \ref{table:query_compare_linf_mnist_cifar} compare the performance of $\ell_\infty$ norm constrained untargeted attacks in terms of success rate and average query count on ImageNet, MNIST, and CIFAR-10 respectively. Bayes attack consistently achieves $2 \times \text{ to } 10 \times$ higher attack success rate while 
requiring $10 \times$ to $20 \times$ fewer queries compared to the current methods across all datasets.

 \begin{table*}[t]
 \centering
 \caption{Results for $\ell_2$ untargeted attacks on ImageNet classifiers with a query budget of $1000$.}
\begin{tabular}{c l c c | c c | c c}
\toprule
\multicolumn{1}{l}{}          & \multicolumn{1}{l}{}  & \multicolumn{2}{c}{$\epsilon = 5.0$}    & \multicolumn{2}{c}{$\epsilon = 10.0$}  & \multicolumn{2}{c}{$\epsilon = 20.0$}  \\

Classifier                    & Method                & success        & avg. query      & success        & avg. query     & success        & avg. query     \\
\midrule
\multirow{6}{*}{ResNet50}     & Boundary Attack       & 8.52           & 655.49          & 15.39          & 577.93         & 26.97          & 538.13         \\
                              & OPT attack            & 7.64           & 777.42          & 15.84          & 737.15         & 32.53          & 757.9          \\
                              & HSJA                  & 6.99           & 904.27          & 14.76          & 887.17         & 28.37          & 876.78         \\
                              & Sign-OPT              & 13.74          & 843.43          & 24.68          & 848.28         & 43.51          & 858.11          \\
                              & Sign-OPT-FFT          & 13.99          & 919.14          & 29.26          & 906.3          & 55.97          & 902.82         \\
                              & \textbf{Bayes attack} & \textbf{20.1}  & \textbf{64.23}  & \textbf{37.15} & \textbf{64.13} & \textbf{66.67} & \textbf{54.97} \\
\midrule
\multirow{6}{*}{Inception-v3} & Boundary Attack       & 0.38           & 998.67          & 6.88           & 588.24         & 14.52          & 531.06         \\
                              & OPT attack            & 3              & 820.79          & 6.88           & 673.71         & 16.27          & 722.75         \\
                              & HSJA                  & 1.63           & 878.54          & 5.51           & 882.48         & 12.89          & 887.8          \\
                              & Sign-OPT              & 7.13           & 839.57          & 14.26          & 831.76         & 22.78          & 837.91         \\
                              & Sign-OPT-FFT          & 7.13           & 929.32          & 15.39          & 924.64         & 34.79          & 917.27         \\
                              & \textbf{Bayes attack} & \textbf{11.39} & \textbf{109.65} & \textbf{22.65} & \textbf{65.66} & \textbf{39.92} & \textbf{68.86} \\
\midrule
\multirow{6}{*}{VGG16-bn}     & Boundary Attack       & 11.23          & 626.33          & 21.27          & 547.64         & 39.37          & 503.22         \\
                              & OPT attack            & 11.09          & 736.58          & 21.79          & 658.9          & 43.86          & 718.68         \\
                              & HSJA                  & 10.3           & 893.2           & 21.53          & 898.15         & 40.82          & 892.56         \\
                              & Sign-OPT              & 19.81          & 841.05          & 35.8           & 843.68         & 60.63          & 857.71         \\
                              & Sign-OPT-FFT          & 21.4           & 916.81          & 42.93          & 910.89         & 70.81          & 907.93         \\
                              & \textbf{Bayes attack} & \textbf{24.04} & \textbf{69.84}  & \textbf{43.46} & \textbf{76.54} & \textbf{71.99} & \textbf{48.95} \\
 \bottomrule
 \end{tabular}
 \label{table:query_compare_l2}
 \end{table*}
 
\subsection{Targeted $\ell_\infty$ Attacks}
Next, we compare the performance of Bayes attack against OPT \citep{cheng2018query} and {Sign-OPT} \citep{signopt} on targeted attack setting where the goal is to produce an adversarial perturbation such that the prediction of the resulting image is the same as the specified target class. We randomly specify the
target label for each example and keep it consistent across different methods. We evaluate the performance of all the methods with a budget of up to $1000$ queries on MNIST and CIFAR-10. We set $\ell_\infty$ perturbation bound $\epsilon$ to $0.1$ for CIFAR-10 and $0.3$ for MNIST. Table  \ref{table:query_compare_linf_mnist_cifar} compares the performance of $\ell_\infty$ norm constrained targeted attacks in terms of success rate and average query count. Bayes attack consistently achieves $10 \times$ higher attack success rate while requiring $7 \times$ fewer queries compared to the current methods on both MNIST and CIFAR-10 datasets. We note that OPT attack is not able to achieve any success in such a low query budget setting and would require a large number of queries to achieve a non-zero success rate.

\subsection{Untargeted $\ell_2$ Attacks}
We also perform untargeted $\ell_2$ attacks with the proposed method. We compare the performance of the {Bayes attack} against Boundary attack \citep{brendel2017decision}, {OPT} attack \citep{cheng2018query}, HopSkipJump attack~(HSJA) \citep{Chen2019HopSkipJumpAttackAQ}, and {Sign-OPT} \citep{signopt}, which is the current state of the art among hard-label black-box attacks within the $\ell_2$ threat model. 
We compare the performance across three different $\ell_2$ perturbation bounds, where we set $\epsilon$ to $5.0$, $10.0$, and $20.0$ respectively. We evaluate the performance of all the methods with a budget of 1000 queries. 



\begin{figure*}[t!]
\centering
\begin{subfigure}[b]{0.75\textwidth}
   \includegraphics[width=\linewidth]{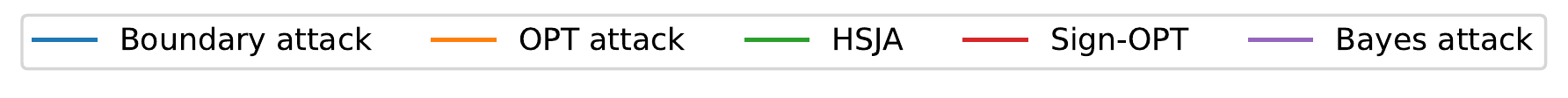}
\end{subfigure}

\begin{subfigure}[b]{0.32\textwidth}
\centering
   \includegraphics[width=0.6\linewidth]{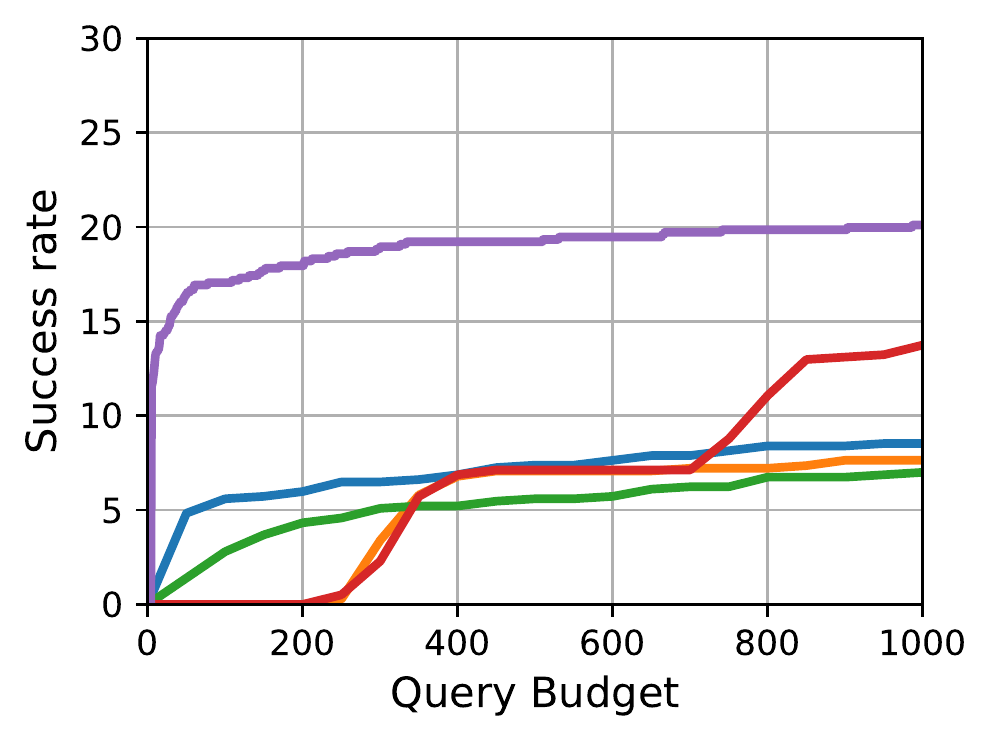}
   \caption{ {ResNet50}, $\epsilon=5.0$}
\end{subfigure}
\centering
\begin{subfigure}[b]{0.32\textwidth}
\centering
   \includegraphics[width=0.6\linewidth]{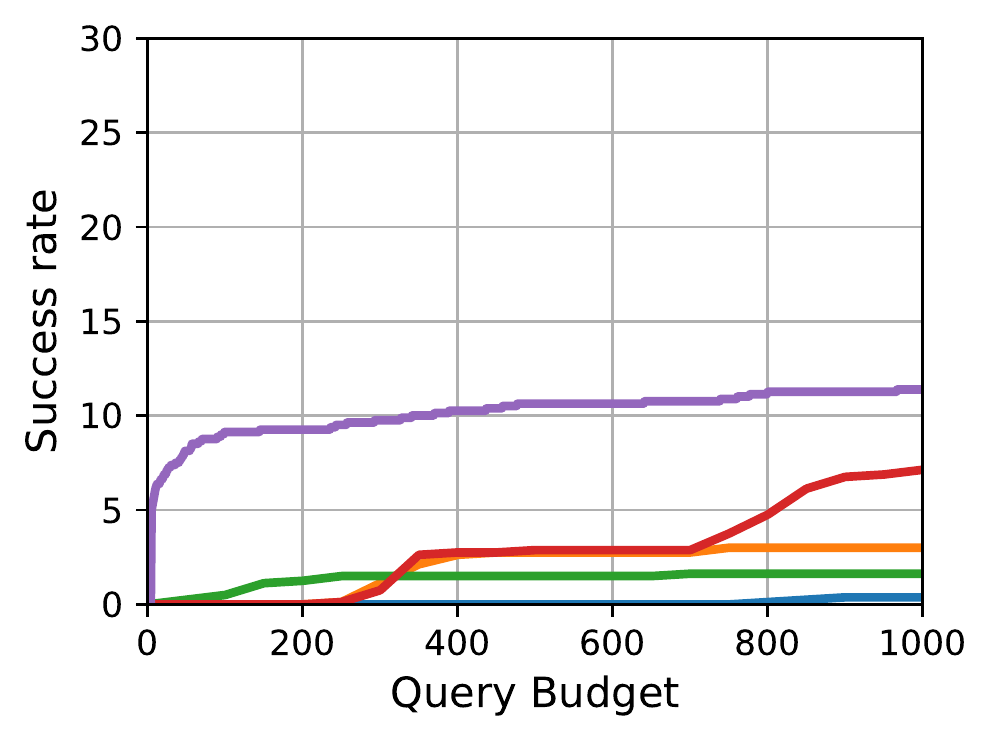}
   \caption{ {Inception-v3}, $\epsilon=5.0$}
\end{subfigure}
\begin{subfigure}[b]{0.32\textwidth}
\centering
   \includegraphics[width=0.6\linewidth]{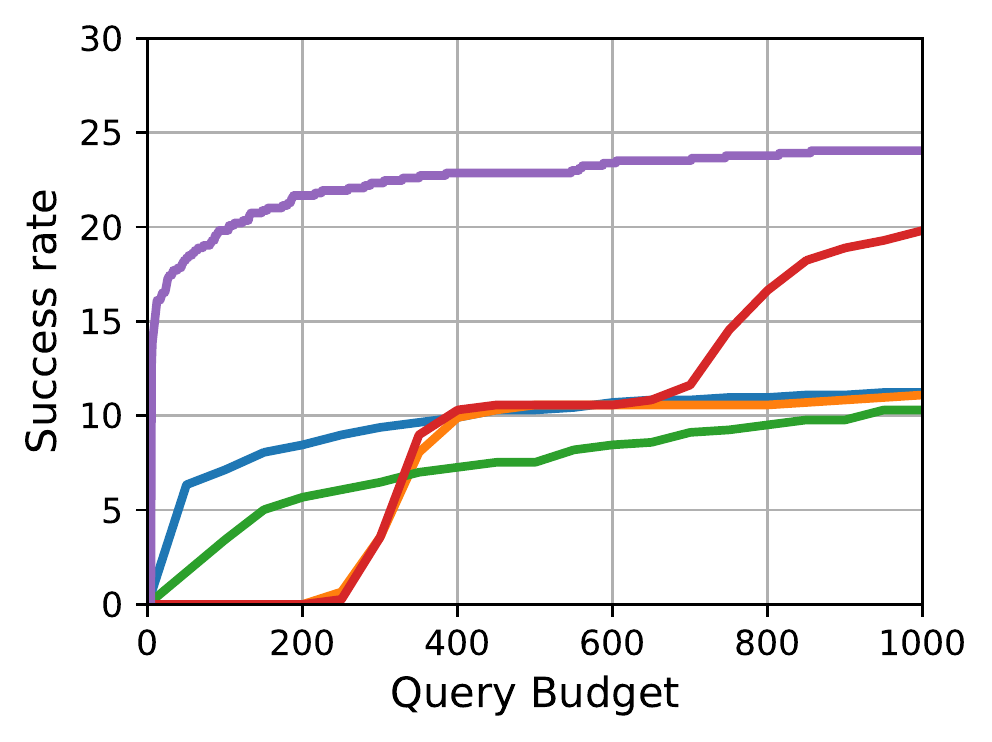}
   \caption{ {VGG16-bn}, $\epsilon=5.0$}
\end{subfigure}
\begin{subfigure}[b]{0.32\textwidth}
\centering
  \includegraphics[width=0.6\linewidth]{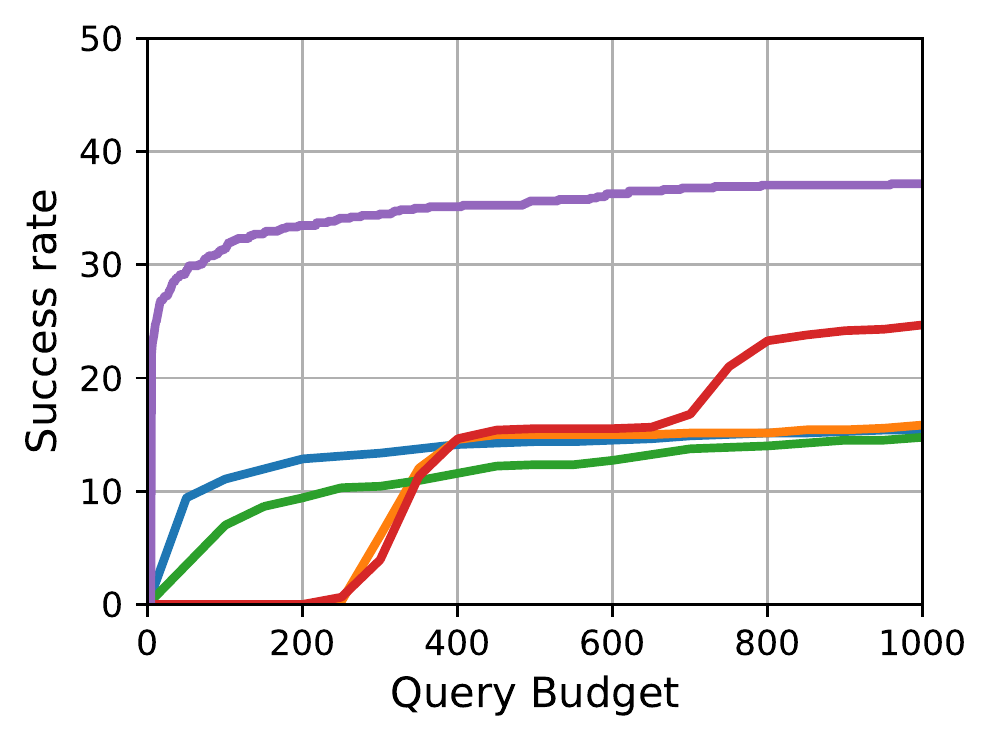}
  \caption{ {ResNet50}, $\epsilon=10.0$}
\end{subfigure}
\begin{subfigure}[b]{0.32\textwidth}
\centering
  \includegraphics[width=0.6\linewidth]{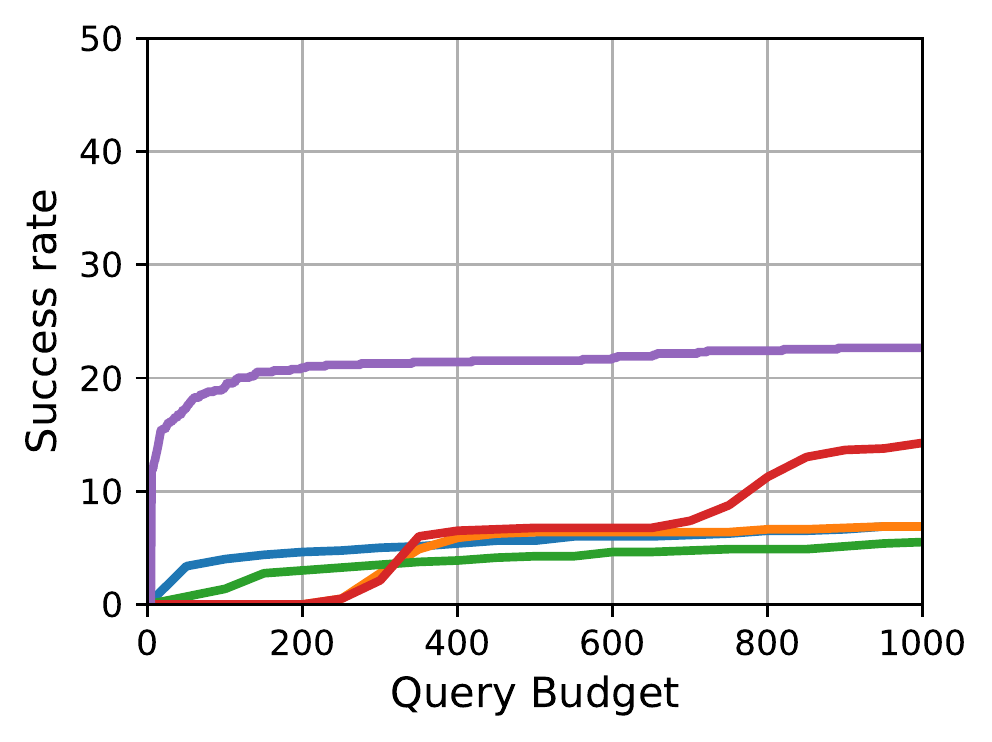}
  \caption{ {Inception-v3}, $\epsilon=10.0$}

\end{subfigure}
\begin{subfigure}[b]{0.32\textwidth}
\centering
  \includegraphics[width=0.6\linewidth]{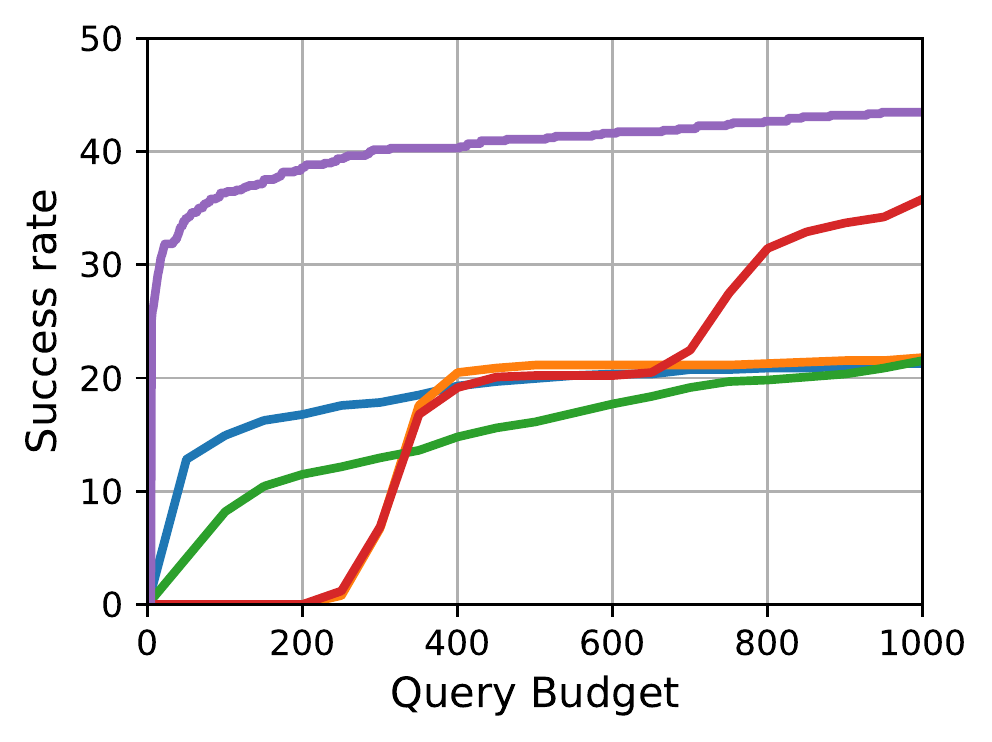}
  \caption{ {VGG16-bn}, $\epsilon=10.0$}
\end{subfigure}
\begin{subfigure}[b]{0.32\textwidth}
\centering
  \includegraphics[width=0.6\linewidth]{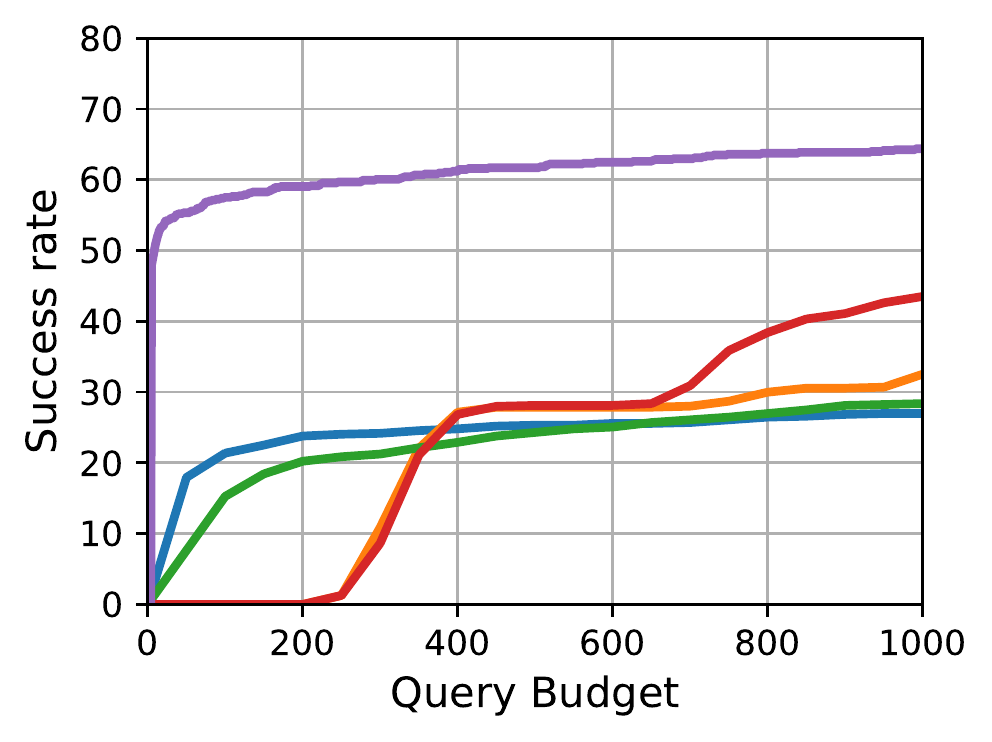}
  \caption{ {ResNet50}, $\epsilon=20.0$}
\end{subfigure}
\begin{subfigure}[b]{0.32\textwidth}
\centering
  \includegraphics[width=0.6\linewidth]{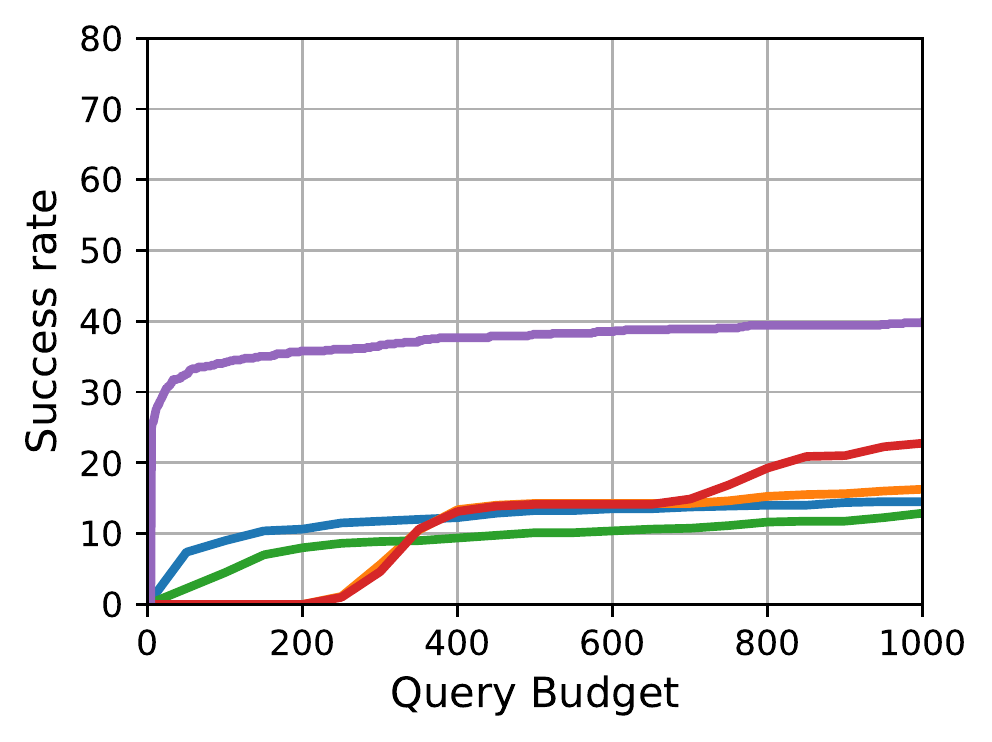}
  \caption{ {Inception-v3}, $\epsilon=20.0$}
\end{subfigure}
\begin{subfigure}[b]{0.32\textwidth}
\centering
  \includegraphics[width=0.6\linewidth]{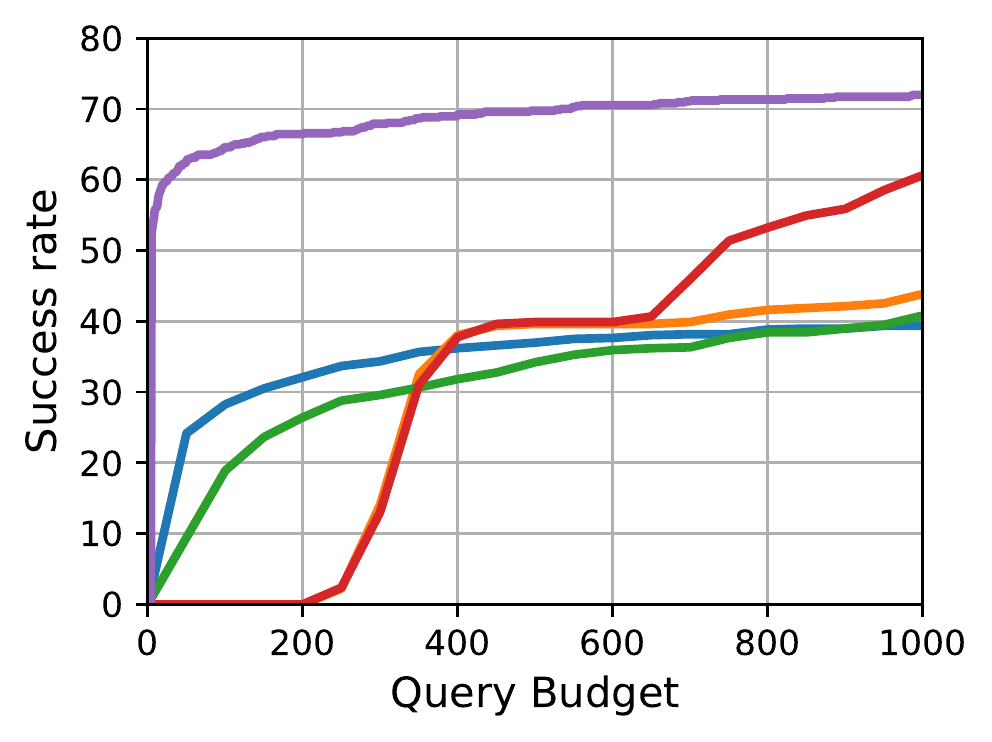}
  \caption{ {VGG16-bn}, $\epsilon=20.0$}
\end{subfigure}
\caption{Performance comparison for $\ell_2$ untargeted attacks on ImageNet classifiers for all methods and $\epsilon$ thresholds.}
\label{fig:pareto_l2_all}
\end{figure*}

Figure \ref{fig:pareto_l2_all} shows the performance of all methods for $\ell_2$ attacks and exhibits the relationship between success rates and the number of queries used for each method on all three ImageNet classifiers and three different $\epsilon$ thresholds for different query budgets. We can see that Bayes attack consistently outperforms the other baseline methods across all query budgets up to $1000$. 
 Table \ref{table:query_compare_l2} compares the success rate and average query count of all methods, models, and $\epsilon$ thresholds for a query budget of $1000$. We can see that  {Bayes attack} consistently outperforms the other baseline methods across all classifiers and epsilons. 
 Bayes attack achieves better success rates and hugely reduces the average query count by up to a factor of $10$ consistently over all the baseline methods. 
 

We perform an ablation study to understand which part of the proposed attack method provides the performance lift. We implement Sign-OPT-FFT, another version of Sign-OPT where we search for the perturbation in a low dimensional subspace defined by FFT basis similar to the proposed method.
 As we can see from Table \ref{table:query_compare_l2} that Sign-OPT-FFT clearly improves the success rate over Sign-OPT but our proposed method still achieves better success rates with reduced average query count. This huge reduction in query counts for finding adversarial perturbations can be contributed to efficiently searching for adversarial perturbations using Bayesian optimization.
 
 
 \subsection{Query Efficiency Comparison}


\begin{figure}[h]

\begin{subfigure}[b]{0.23\textwidth}
\centering
   \includegraphics[width=.8\linewidth]{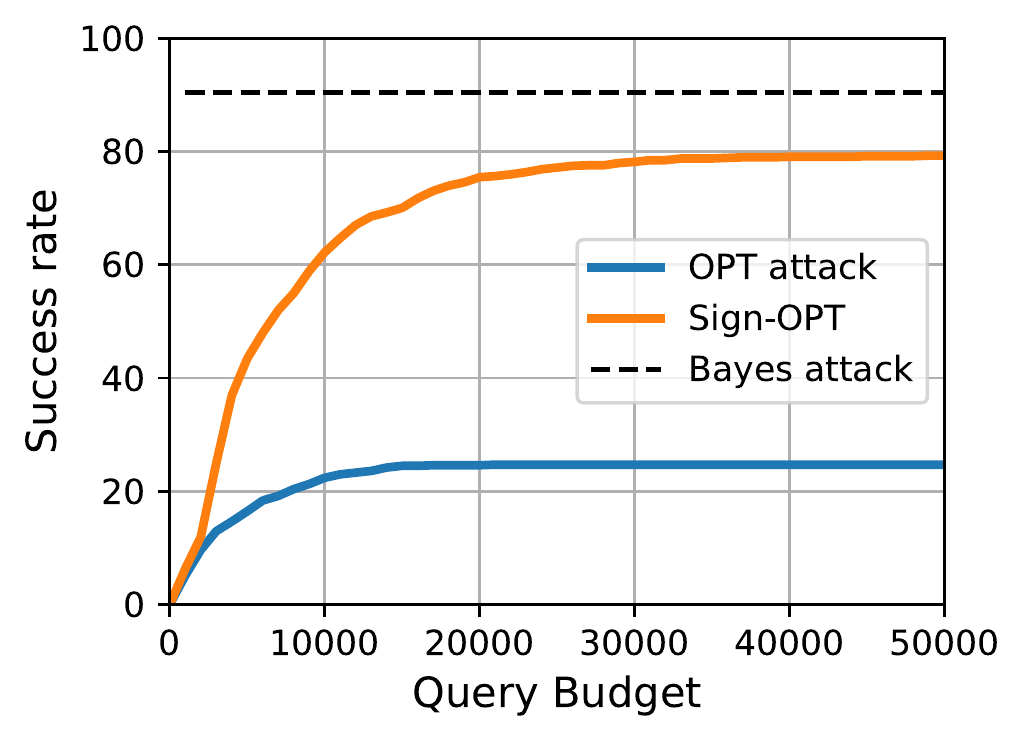}
   \caption{MNIST}
    \label{fig:pareto_mnist}
\end{subfigure}
\begin{subfigure}[b]{0.23\textwidth}
\centering
   \includegraphics[width=.8\linewidth]{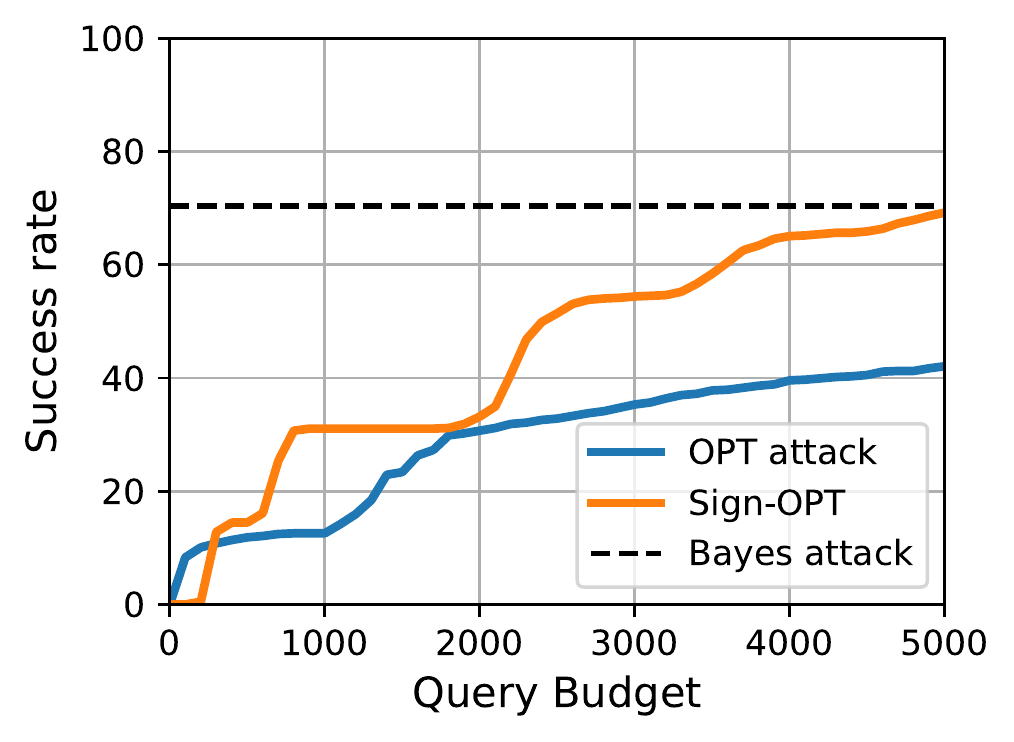}
   \caption{CIFAR-10}
   \label{fig:pareto_cifar}
\end{subfigure}
\caption{\small{Query efficiency comparison on MNIST and CIFAR-10.}}
\end{figure}

We compare the query efficiency of the competing methods on MNIST and CIFAR-10 for $\ell_\infty$ attacks. We run the competing methods until they achieve the same best attack success rate as Bayes attack or exceed a much higher query limit ($50000$ for MNIST and $5000$ for CIFAR-10). We use the same setting as described in Section 5.2. Figures \ref{fig:pareto_mnist} and \ref{fig:pareto_cifar} compare the performance on MNIST and CIFAR-10 respectively. For MNIST, we can see that even with $50 \times$ higher query budget than the Bayes attack, competing methods are not able to achieve a similar attack success rate. For CIFAR-10, Sign-OPT requires $5 \times$ higher queries to achieve the same success rate as Bayes attack. 

Finally, based on the experimental evidence we note that given a query budget of 50,000 to 100,000 or more, it is possible for the current approaches to achieve an attack success rate close to that of the Bayes attack (with a query budget of 1000). In constrained or low query budgets, the Bayes attack hugely outperforms the current state-of-the-art approaches.


\subsection{Low Dimension Subspaces}
\label{sec:basis}
In this section, we perform ablation studies on ImageNet by exploring different low dimensional structured subspaces.
In the case of the $\ell_2$ threat model, we utilize the low dimensional subspace generated by the low-frequency sine and cosine FFT basis vectors. We can also consider the cosine FFT basis or sine FFT basis separately by using only the real components or imaginary components of the frequency domain representation. 

We compare the attacks generated in the low dimension subspace created using cosine and sine FFT basis vectors separately and together. For this experiment, we perform hard label black-box attacks on ResNet50 trained on ImageNet with $\ell_2$ perturbation set to 20.0. We maintain a query budget of 1000 across the experiments. For a fair comparison, we keep the size of low dimension subspace almost the same across the experiments, i.e. $3 \times 18 \times 18$ for cosine and sine FFT basis separately and $3 \times 12 \times 12 \times 2$ when considering the complete FFT basis. We also compare with a random set of vectors sampled from the standard normal distribution. We keep the size of vectors sampled from normal distribution same as $3 \times 18 \times 18$.

 \begin{table}[t]
\centering
\caption{Performance comparison of FFT basis vectors and random vectors sampled from the standard normal distribution for $\ell_2$ attack with $\epsilon=20.0$ on ResNet50.}
\begin{tabular}{c c  c } 
 \toprule
 {\bf Basis } & {\bf Success} & {\bf Avg Queries} \\
 \midrule
 {Cosine FFT} & $  {64.38}	\% $ & $	{54.25}	 $ \\
 {Sine FFT}	    & $	{63.74}	\% $ & $	{45.72}	$ \\
 {Cosine and sine FFT} & $	{66.67}	\% $ & $	{54.97}	$\\
 {Standard Normal}  & $	{33.33}	\% $ & $	{48.25}	$ \\
 \bottomrule
 \end{tabular}
 
 \label{table:query_compare_fft}
 \end{table}
 
 
 Table \ref{table:query_compare_fft} compares the performance of basis vectors in terms of attack success rate and average query count. We observe that the low-frequency FFT basis vectors enhanced the attack accuracy significantly as compared to the random set of vectors generated from standard normal distribution. On the other hand among low-frequency components, sine and cosine FFT basis vectors together provide a slight gain in attack success rate as compared to using them separately. 
 
\begin{table}[t]
\centering
\caption{Comparing inverse fast Fourier transform (IFFT) and nearest neighbor interpolation (NNI) for $\ell_2$ and $\ell_\infty$ attack on ResNet50.}
\begin{tabular}{c c  c c } 
 \toprule
{{ \bf Attack}} & {\bf Mapping } & {\bf Success} & {\bf Avg } \\
{\bf Type} & {\bf Technique} & {\bf Rate} &{\bf Queries}\\
 \midrule
\multirow{2}{*}{$\ell_\infty, \epsilon = 0.05$ } 
 
&	{IFFT} & $  {59.16}	\% $ & $	{55.72}	 $ \\
&	{NNI}	& $	{67.48}	\% $ & $	{45.94}	$ \\
\midrule
\multirow{2}{*}{$\ell_2, \epsilon = 20.0$}
&	{IFFT} & $	{66.67}	\% $ & $	{54.97}	 $ \\
&	{NNI}	& $	{59.54}	\% $ & $	{50.71} $  \\

 \bottomrule
 \end{tabular}
 
 \label{table:query_compare_upsampling}
 \end{table}
 
\subsection{Mapping Techniques}
The proposed method requires a low dimensional subspace for efficiently searching the perturbation and a mapping technique for transforming the perturbation learned in the low dimension to the full input space.  
We use FFT basis vectors for learning perturbation for $\ell_2$ threat model while nearest neighbor interpolation for $\ell_\infty$ threat model. Here, we compare both the methods on $\ell_2$ as well as $\ell_\infty$ threat models.

We compare both the mapping techniques on attacking ResNet50 trained on ImageNet with $\ell_\infty$ and $\ell_2$ perturbation set to $0.05$ and $20.0$, respectively. We maintain a query budget of 1000 across the experiments. For a fair comparison, we keep the size of low dimension subspace the same across the experiments, i.e. $3 \times 18 \times 18$. 

Table \ref{table:query_compare_upsampling} shows the performance of both the mapping techniques on $\ell_\infty$ and $\ell_2$ threat models. FFT basis vectors perform better than nearest neighbor interpolation in the $\ell_2$ threat model, while nearest neighbor performs better for the $\ell_\infty$ threat model. This could be because of the isometric property of the mapping technique (inverse FFT transformation and nearest neighbor interpolation lead to equivalent $\ell_2$ and $\ell_\infty$ norms respectively) with respect to the $\ell_p$-norm.

%% file: conclusions.tex
\section{Conclusions}

We consider the problem of hard-label black-box adversarial attacks in low query budget regimes  which is an important practical consideration.
To efficiently generate adversarial attacks with higher success rates and fewer queries, we define two low dimension structured subspaces favorable for $\ell_2$ and $\ell_\infty$ norm constrained hard-label black-box attacks. Our proposed method uses Bayesian optimization for finding adversarial perturbations in low dimension subspace and maps it back to the original input space to obtain the final perturbation. We successfully demonstrate the efficacy of our method in attacking multiple deep learning architectures for high dimensional inputs in both untargeted and targeted attack settings, and $\ell_\infty$ and $\ell_2$ threat models. 
Our work opens avenues regarding applying BO for black-box adversarial attacks in high dimensional settings with low query budgets.